\documentclass{article}

\usepackage[utf8]{inputenc}
\usepackage[T1]{fontenc}
\usepackage{graphicx}
\usepackage{amsmath,amsthm,amssymb,amsfonts,mathtools}
\usepackage{hyperref}
\usepackage{cleveref}
\usepackage{algorithm2e}
\usepackage{framed}
\usepackage{url}
\usepackage{booktabs}
\usepackage{nicefrac}
\usepackage{microtype}
\usepackage{color}

\SetKwProg{when}{whenever}{~do}{end}
\SetKwInput{init}{initialize}
\SetAlFnt{\sffamily}

\SetKwSty{sftmp}
\SetArgSty{sftmp}

\crefname{algocf}{alg.}{algs.}
\Crefname{algocf}{Algorithm}{Algorithms}

\newcommand\mnist{\textsc{mnist} }

\title{Deep counter networks \\ for asynchronous event-based processing}

\author{
	Jonathan Binas$^*$, Giacomo Indiveri, and Michael Pfeiffer\\
	Institute of Neuroinformatics\\
	University of Zurich and ETH Zurich, Switzerland\\
  	\texttt{$^*$jbinas@ini.ethz.ch} \\
}

\begin{document}
\maketitle

\begin{abstract}
  Despite their advantages in terms of computational resources, latency, and power consumption, event-based implementations of neural networks have not been able to achieve the same performance figures as their equivalent state-of-the-art deep network models. We propose counter neurons as minimal spiking neuron models which only require addition and comparison operations, thus avoiding costly multiplications. We show how inference carried out in deep counter networks converges to the same accuracy levels as are achieved with state-of-the-art conventional networks. As their event-based style of computation leads to reduced latency and sparse updates,  counter networks are ideally suited for efficient compact and low-power hardware implementation. We present theory and training methods for counter networks, and demonstrate on the \mnist benchmark that counter networks converge quickly, both in terms of time and number of operations required, to state-of-the-art classification accuracy.
\end{abstract}

 \section{Introduction}

 Despite the remarkable success of deep neural networks~\cite{lecun2015deep} in areas such as computer vision~\cite{krizhevsky2012imagenet,he2015delving} or speech recognition~\cite{hinton2012deep}, biological neural systems clearly outshine their artificial counterparts in terms of compactness, speed and energy consumption.
 One putative reason for such efficiency may lie in the way signals are represented and transmitted in animal brains: data is transmitted sparsely and asynchronously, in small packets by means of spikes.
 This is in stark contrast with the frame-based approach of classical neural networks, which always compute the complete output of one layer synchronously, before passing it on to the next layer.
 Indeed, spike-based processing allows for more efficient utilization of communication channels and computing resources, and can lead to speedups in processing \cite{neil2016learning}.
 These advantages have sparked interest in dedicated spiking neural network electronic devices based on such event-driven processing schemes~\cite{merolla2014million,indiveri2015neuromorphic}.
 Achieving the same accuracy as state-of-the-art deep learning models with event-based updates has remained challenging, but recently a number of methods have been proposed which convert a previously trained conventional analog neural network (ANN) into a spiking one~\cite{cao2015spiking,diehl2015fast,hunsberger2015spiking,Esser20092016}. The principle of such conversion techniques is to approximate the continuous-valued activations of ANN units by the spike rates of event-based neurons. Although successful on several classical benchmark tasks, all these methods suffer from approximation errors, and typically require a multitude of spikes to represent a single continuous value, thereby losing some of the advantages of event-based computation.

 In this work, we propose a set of minimalistic event-based asynchronous neural network models, which process input data streams continuously as they arrive, and which are formally equivalent to conventional frame-based models.
 This class of models can be used to build highly efficient event-based processing systems, potentially in conjunction with event-based sensors~\cite{delbruck2010activity,posch2014retinomorphic}. The resulting systems process the stream of incoming data in real time, and yield first predictions of the output typically already after a few data packets have been received, and well before the full input pattern has been presented to the network.

Here we demonstrate how the computation carried out by counter networks exactly reproduces the computation done in the conventional frame-based network. This allows maintaining the high accuracy of deep neural networks, but does so with a power and resource efficient representation and communication scheme.
Specifically we show how, as a consequence of the event-driven style of processing, the resulting networks do not require computationally expensive multiplication operations. We initially demonstrate the principle for networks with binary activations, and then extend the model to non-binary activations. The performance of our novel models is evaluated on the \mnist dataset.
 The numerical results indicate that counter networks require fewer operations than previous approaches to process a given input, while enabling state-of-the-art classification accuracy.

 \section{Counter neural networks}

 Multiplications are the most expensive operations when using conventional neural networks for inference on digital hardware. It is therefore desirable to reduce the number of required multiplications to a minimum.
 In this section, we introduce an event-based neuron model, which only makes use of addition operations, counter variables, and simple comparisons, all of which can be implemented very efficiently in simple digital electronic circuits.

 \subsection{Multiplication-free networks}

 Previous work has shown how frame-based neural networks can be implemented using additions only, by either restricting all weights to binary (or ternary) values~\cite{hwang2014fixed,lin2015neural,courbariaux2015binaryconnect}, or by using binary activations~\cite{courbariaux2016binarynet}. The binary variable (weight or activation) then represents an indicator function, and all neural activations can be computed by simply adding up the selected weights or activations.
 To introduce our event-based model in its most basic form, we first investigate the case where neurons have binary activations, such that the output of all neurons $y_i^{\scriptscriptstyle (k)}$ in layer $k$ is given by
 \begin{align}
	 \mathbf{y}^{(k)} = \sigma\left(\mathbf{W}^{(k)} \mathbf{y}^{(k-1)} - \mathbf{\theta}^{(k)}\right),
	 \label{eqn:neuron}
 \end{align}
 where $\mathbf{W}$ is the weight matrix, $\mathbf{\theta}$ are threshold values corresponding to bias terms, and
 \begin{align}
	 \sigma(x)=\begin{cases}
		 1, & \text{if $x>0$}\,, \\
		 0, & \text{otherwise}.
	 \end{cases}
	 \label{eqn:f_bin}
 \end{align}
 Thus, the output of a neuron is 1 if its net input is greater than its threshold value $\theta$ and 0 otherwise.
 While this model does not pose any constraints on the weights and thresholds, we use low precision integer values in all experiments to keep the computational cost low and allow for highly efficient digital implementations.
 We consider here multi-layer networks trained through stochastic gradient descent using the backpropagation algorithm \cite{rumelhart1986learning}.
 Since the error gradient is zero almost everywhere in the discrete network given by \cref{eqn:neuron,eqn:f_bin}, we replace the binarization function $\sigma$ by a logistic sigmoid function $\tilde{\sigma}$ in the backward pass,
 \begin{align}
	 \tilde{\sigma}(x) = \frac{1}{1 + \exp{(-\lambda x)}}\,,
	 \label{eqn:sigmoid}
 \end{align}
 where $\lambda$ is a scaling factor. Furthermore, during training, we keep copies of the high-resolution weights and activations, and use them to compute the gradient in the backward pass, as proposed by \cite{courbariaux2015binaryconnect,stromatias2015robustness}.
 In addition to the activations and network parameters, the inputs to the network are binarized by scaling them to lie in the range $[0,1]$ and rounding them to the nearest integer.

 \subsection{Lossless event-based implementation through counter neurons}

 The multiplication-free network proposed above can directly be turned into an asynchronous event-based neural network by turning every unit of the ANN into a \emph{counter neuron}, which we describe below. The weights and biases obtained through the training procedure above can be used in the event-based network without further conversion.

 \begin{figure}[t]
     \centering
     \includegraphics[scale=0.9]{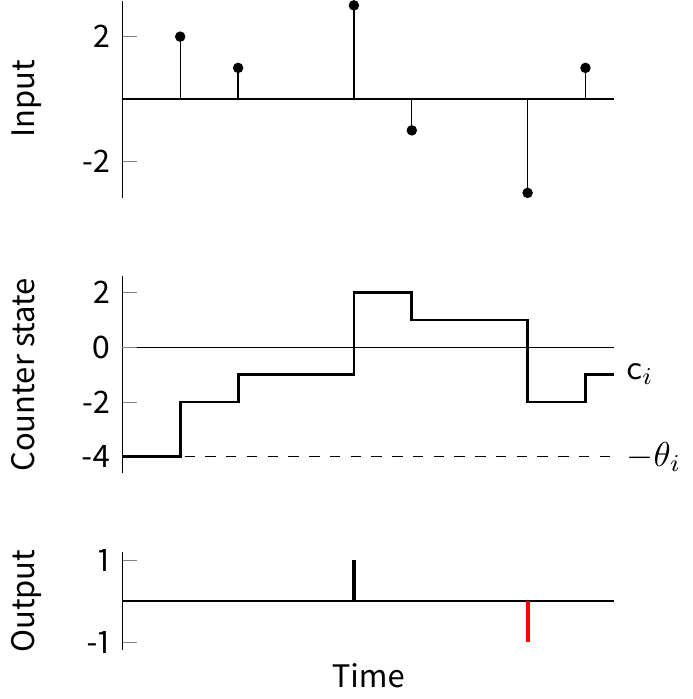}
     \caption{Illustration of the basic counter neuron model. The internal counter variable $c_i$ is initialized at the negative threshold level $-\theta_i$ and accumulates inputs provided through events at discrete time steps. Whenever the counter changes its sign, i.e. when the accumulated input becomes greater or smaller than the threshold level $\theta_i$, an event is emitted. The event is positive ($+1$) if the sign changes from negative to positive, and negative ($-1$) otherwise.
     }
     \label{fig:illust}
 \end{figure}

 \begin{algorithm}[t]
 \centering
 \begin{minipage}[]{0.75\textwidth}
 \begin{leftbar}
	 \DontPrintSemicolon
	 \init{ $c_i \leftarrow -\theta_i$\; }
	 \when{
 neuron $i$ receives an input event $\mathrm{inp}_i=\mathrm{sign}_{j^*} w_{ij^*}$ from a neuron $j^*$ or a sum of simultaneous input events $\mathrm{inp}_i=\sum_{j \in J^*} \mathrm{sign}_{j} w_{ij}$ from a set of neurons $J^*$ }
	 {
		 \BlankLine
		 $c_i^\mathrm{prev} \leftarrow c_i$\;
		 $c_i \leftarrow c_i + \mathrm{inp}_i$\;
		 \BlankLine
		 \If { $c_i^\mathrm{prev} \leq 0$ and $c_i > 0$ }{
			 emit $+1$\;
		 }
		 \BlankLine
		 \If { $c_i^\mathrm{prev} > 0$ and $c_i \leq 0$ }{
			 emit $-1$\;
		 }
	 }
 \end{leftbar}
 \end{minipage} \\[1em]
 \caption{Basic counter neuron implementation.}
 \label{alg:neuron}
 \end{algorithm}

 Each \emph{counter neuron} is defined by an internal counter variable $c$, which is updated whenever the neuron receives positive or negative inputs in the form of binary events (or spikes). The neuron operation is illustrated in \cref{fig:illust} and is described by \cref{alg:neuron}.
 A counter neuron essentially counts, or adds up, all the inputs it receives.
 Whenever this sum crosses the threshold $\theta$, a binary event $\pm1$ is emitted. The value of the event depends on the direction in which the threshold was crossed, i.e. a positive event is emitted when the threshold is crossed from below, and a negative event when $c$ falls below the threshold.
 Whenever neuron $j$ emits an event the quantity $\pm w_{ij}$ is provided as input to neuron $i$ of the next layer, with the sign determined by the output of neuron $j$.
 Thus, the neurons themselves do not continuously provide output signals, which makes information transmission and computation in the network very sparse.
 The input to the network is also provided in event-based form as a stream of binary events (or spikes), i.e. the network at discrete points in time receives a list of indices of one or more pixels, indicating that these pixels are active.
 Specifically, a binary input image (or other data) is streamed to the network pixel-by-pixel in an asynchronous manner, whereby the order and exact timing of the pixels does not matter.
In the following we will show analytically that an event-based network based on counter neurons produces exactly the same output as its frame-based counterpart.

 \begin{proof}[Proof of the equivalence]
 To prove the equivalence of the frame-based and the event-based model we have to show that the outputs of individual neurons are the same in both settings.
 In the following, we assume $\theta_k > 0\,\forall k$ without loss of generality.
 Let an event-based network be initialized at time $t=0$, such that all $c_k(0)=-\theta_k$.
 Within the time interval $[0,T]$, neuron $i$ of layer $m$ in the event-based network receives a sequence of $N$ inputs, $(w_{ij_1}^{\scriptscriptstyle (m)} s_1,\ldots,w_{ij_N}^{\scriptscriptstyle (m)} s_N)$ from a set of source neurons $j_1,\ldots,j_N$ at times $t_1,\ldots,t_N$, where $s_k$ is the sign of the $k$th event, and $0 \leq t_k \leq T, \forall k$.
 It follows from \cref{alg:neuron} that the value of the counter variable $c_i^{\scriptscriptstyle (m)}$ at time $T$ is
 \begin{align}
      c_i^{(m)}(T) 
      = c_i^{(m)}(t_N)
      = \textstyle\sum_{k=1,\ldots,N} w_{ij_k}^{(m)} s_k - \theta_i^{(m)}\,,
     \label{eqn:equiv:cvar}
 \end{align}
 as it simply sums up the inputs.
 The sign of $c_i^{\scriptscriptstyle (m)}$ might change several times during the presentation of the input, and trigger the emission of a positive or negative event at every zero-crossing.
 Since $c_i^{\scriptscriptstyle (m)}$ is initialized at $-\theta_i^{\scriptscriptstyle (m)} < 0$, there are $2n + \sigma(c_i^{\scriptscriptstyle (m)}(t_N))$ sign changes in total, where $n$ is a non-negative integer, and thus the total input communicated to neuron $k$ of the next layer is
 \begin{align}
	 \sigma\left(c_i^{(m)}(t_N)\right) w_{ki}^{(m+1)}
	 = \sigma\left(\textstyle\sum_{k=1}^{N} w_{ij_k}^{(m)} s_k - \theta_i^{(m)} \right) w_{ki}^{(m+1)}
	 \eqqcolon \hat{y}_i^{(m)} w_{ki}^{(m+1)}\,,
	 \label{eqn:equiv}
 \end{align}
 as the $2n$ sign changes cancel out. On the other hand, recursively applying \cref{eqn:equiv} leads to
 \begin{align}
     \hat{y}_i^{(m)}
     = \sigma\left(\textstyle\sum_j w_{ij}^{(m)} \sigma(c_j^{(m-1)}(t_N)) - \theta_i^{(m)} \right)
     = \sigma\left(\textstyle\sum_j w_{ij}^{(m)} \hat{y}_j^{(m-1)} - \theta_i^{(m)} \right)\,.
     \label{eqn:equiv:iter}
 \end{align}
 Since $\hat{y}_k^{\scriptscriptstyle (0)} = y_k^{\scriptscriptstyle (0)}$ for the input layer, the equivalence must also hold for all higher layers, according to \cref{eqn:equiv:iter}.
\end{proof}

\begin{figure}[t]
	 \centering
	 \includegraphics[scale=0.9]{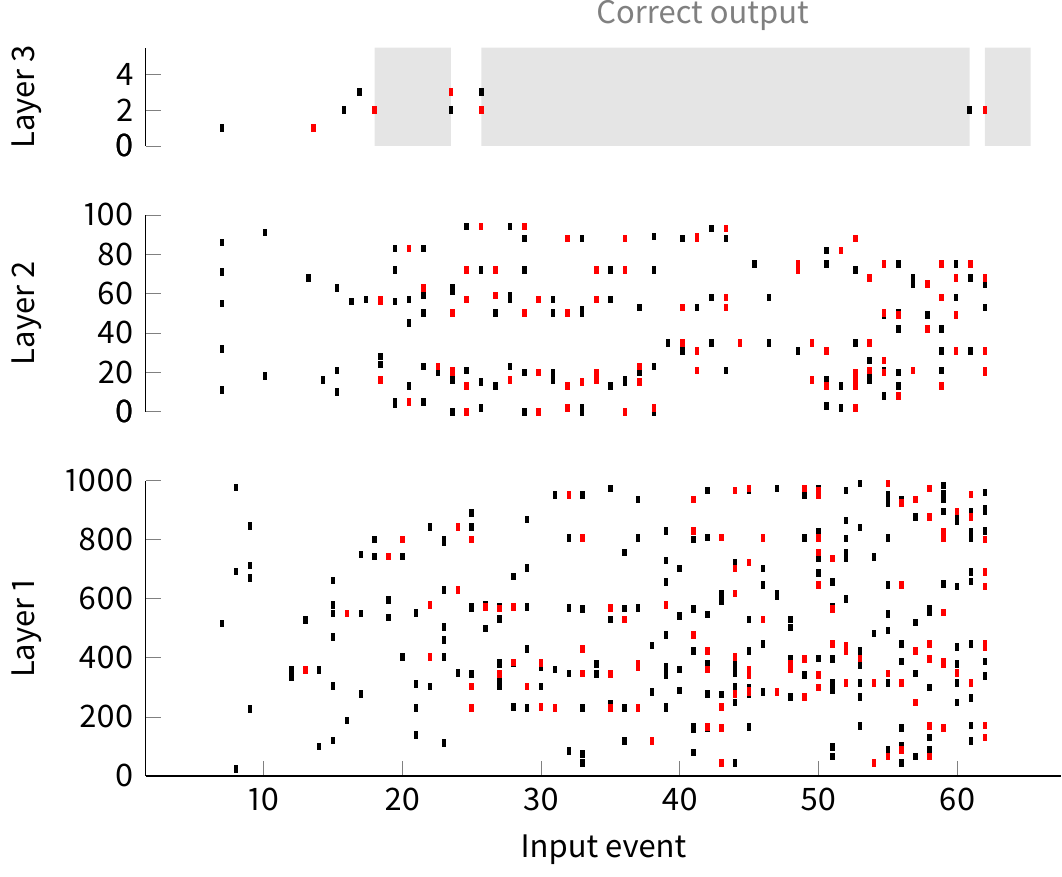}
	 \caption{Example run of the event-based system.
         A single input pattern of the \mnist test set is presented over time (one pixel per timestep) to a trained counter network. Positive events are displayed in black, negative events in red. The correct output (class 3) for this input pattern can be read out long before the whole input has been presented. Shaded regions indicate that the output of the network is correct during these periods, i.e. only the output neuron for class 3 has produced more positive than negative output spikes.}
	 \label{fig:spiking}
 \end{figure}

With the notion of equivalence, it is clear that the event-based network, if configured with the parameters obtained for the frame-based network, is able to exactly reproduce the output of the latter.
Unlike in previous work~\cite{cao2015spiking,diehl2015fast}, the resulting event-based network is guaranteed to provide the same result as the `ideal' frame-based implementation.
 Thereby, the respective output value can be obtained by adding up the events emitted by the output layer.
 Technically, the equivalence holds only in the case where the full stream of input events has been presented to the network, and propagated through all layers.
 In practice, however, a few input events are often enough to activate the right neurons and produce the correct output long before the full set of input events has been presented.
 As a consequence, on average far fewer operations than in the frame-based model are required to compute the correct output (see \cref{fig:spiking} for an example).

 \subsection{Extension to non-binary inputs}

 The constraint that the input patterns are binary, and each input unit either produces an event or not, can be safely relaxed to integer-valued inputs without further modifications of the model.
 The framework thus supports finer grained input scales, which is important e.g. for input images using multiple gray-levels or RGB values to encode different colors.
 The simple modification is that each individual input unit produces over time a number of events that corresponds to the encoded integer value.
 While such integer-valued inputs would require multiplication in the frame-based network with non-binary (or non-ternary) weights, the event-based network remains free of multiplication operations, as the instantaneous output of any input unit is still binary.

 \subsection{Extended counter neuron network with non-binary activations}

 \begin{algorithm}[t]
     \centering
     \begin{minipage}{0.75\textwidth}
	 \begin{leftbar}
	     \DontPrintSemicolon
	     \init{ 
		 $z_i \leftarrow 0$;~
		 $c_i \leftarrow -\theta_i$\;
	     }
	     \BlankLine
	     \when{ \it there is an input event $\mathrm{inp}_i=\mathrm{sign}_{j^*} w_{ij^*}$ from a neuron $j^*$ or a sum of simultaneous input events $\textrm{inp}_i = \sum_{j \in J^*} \mathrm{sign}_{j} w_{ij}$ from a set of neurons $J^*$ }
	     {
		 \BlankLine
		 $c_i \leftarrow c_i + \mathrm{inp}_i$\;
	     \BlankLine
	     \While { $c_i \geq \lambda$ }{
		 emit $+1$\;
		 $c_i \leftarrow c_i - \lambda$\;
		 $z_i \leftarrow z_i + 1$\;
	     }
	     \BlankLine
	     \While { $z_i > 0$ and $c_i < 0$ }{
		 emit $-1$\;
		 $c_i \leftarrow c_i + \lambda$\;
		 $z_i \leftarrow z_i - 1$\;
	     }
	     }
	 \end{leftbar}
     \end{minipage} \\[1em]
     \caption{Extended counter neuron implementation based on the discretized ReLU activation. The parameter $\theta_i$ represents the neuron's threshold. The scaling factor $\lambda$ allows adjusting the step size at which the neuron emits events, i.e. how much more input is required to trigger the next event.}
     \label{alg:ext_neuron}
 \end{algorithm}

 \begin{figure}[t]
     \centering
     \includegraphics[scale=0.9]{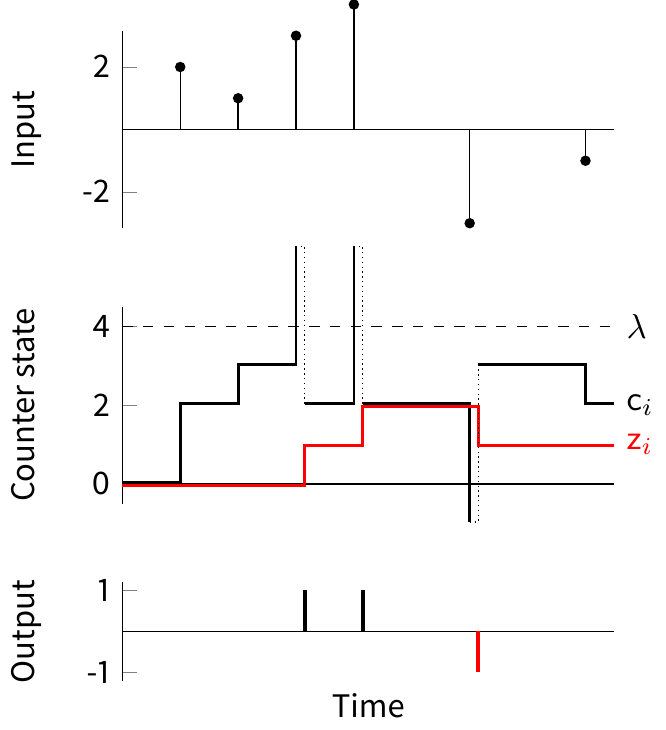}
 \caption{Illustration of the extended counter neuron model. The internal counter variable $c_i$ accumulates input events at discrete time steps. Whenever the variable crosses the threshold level $\lambda$, a positive event ($+1$) is emitted and the second counter variable $z_i$ is incremented by 1. On the other hand, if $z_i > 0$ and $c_i$ becomes negative, a negative event ($-1$) is emitted and $z_i$ is decremented. Whenever an event is emitted, $c_i$ `wraps around' to stay in the range $(0,\lambda]$. Note that we assumed $\theta_i=0$ for this illustration.
 }
     \label{fig:ext-illust}
 \end{figure}

 Using binary outputs for neurons might be a disadvantage, because due to the the limited output bandwidth of individual neurons it might be necessary to use more neurons in hidden layers to obtain the same accuracy as a  non-binary network.
 The counter network model can easily be extended to non-binary activations, without requiring multiplication in the event-based implementation.
 In order to train a network based on neurons featuring multiple, discrete levels of activation, we use a discretized version of the ReLU activation function during training:
 \begin{align}
     f(x) = \rho\left(\left\lfloor\frac{x+\epsilon}{\lambda}\right\rfloor\right),
     \label{eqn:drelu}
 \end{align}
 where $\epsilon\ll 1$ is a small, positive constant, $\lambda \in \mathbb{Z}^+$ is a scaling factor, and $\rho$ is the typical ReLU half-wave rectification,
 \begin{align}
	 \rho(x)=\begin{cases}
		 x, & \text{if $x>0$}\,, \\
		 0, & \text{otherwise}.
	 \end{cases}
	 \label{eqn:f_relu}
 \end{align}

 As in the binary case, the discrete activation is used during training in the forward pass, and a continuous approximation in the backward pass. Specifically, $f$ can be approximated by a shifted and scaled ReLU,
 \begin{align}
     \tilde{f}(x) = \rho\left(\frac{x}{\lambda} - \frac{1}{2}\right),
     \label{eqn:appr_drelu}
 \end{align}
 in the backward pass.
 The learned parameters can then again be directly transferred to configure a network of event-based neurons without further conversion of weights or biases. The dynamics of this network are illustrated in \cref{fig:ext-illust}, and described in \cref{alg:ext_neuron}.
 The equivalence of the frame-based and the event-based implementation can be proven similarly to the binary case. A sketch of the proof is outlined below:

 \begin{proof}[Proof of the equivalence]
 From \cref{alg:ext_neuron} it can be seen that after a neuron has processed a number of input events, its internal variable $z_i$ has the value $\rho( \lfloor (x_i - \theta_i + \epsilon)/\lambda \rfloor )$, where $x_i$ is the accumulated input provided over time.
 On the other hand, the value of $z_i$ changes only when an event is emitted, and its value corresponds to the number of positive events emitted, minus the number of negative events emitted.
 Thus, the accumulated output communicated by the neuron corresponds precisely to $z_i$, and thereby to the output of the frame-based model given by \cref{eqn:drelu}, since $x_i$ corresponds to the total input provided by neurons from the previous layer, $x_i = \sum_j w_{ij} y_j$.
 \end{proof}

 The discretized ReLU offers a wider range of values than the binary activation, and therefore allows for a more fine-grained response, thereby facilitating training.
 On the other hand, using non-binary outputs might lead to larger output delays compared to the binary case, as the trained neurons might now require a multitude of events from individual neurons to arrive at the correct output.

 \section{Results}

 Various networks were trained on the \mnist dataset of handwritten digits to demonstrate competitive classification accuracy.
 In particular, we evaluated fully connected networks (FCNs) of three hidden layers (784-1000-1000-1000-10) and five hidden layers (784-750-750-750-750-750-10) to investigate how the depth of the network affects the processing speed and efficiency. In addition we trained convolutional networks (CNNs) with two (784-12c5-12c7-10) and four (784-12c3-12c5-12c7-12c9-10) hidden layers.
 The network dimensions were chosen such that the number of parameters remains roughly the same in the shallower and deeper networks ($\approx$2.8 mio. parameters for the FCNs, and $\approx$50k for the CNNs.)

 \subsection{Training details}

 The networks were trained through stochastic gradient descent using the Adam method \cite{kingma2014adam}.
 The gradients for the backward pass were computed using the continuous approximations described by \cref{eqn:sigmoid,eqn:appr_drelu}.
 All parameters were restricted to 8-bit integer precision in the forward pass, and floating point representations were used only in the backward pass, as suggested by \cite{courbariaux2015binaryconnect, stromatias2015robustness}.
 The biases $\theta$ were restricted to non-negative values through an additional constraint in the objective function, otherwise
 categorical cross-entropy was used as the loss function.
 The learning rate was set to 0.01 for the CNNs and to 0.005 for the FCNs.
 The \mnist dataset was split into a training set of 50000 samples, a validation set of 10000 samples, and a test set of 10000 samples, and
 a batch size of 200 samples was used during training.
 The networks were trained until a classification error of $\approx 1.5\%$ on the validation set was obtained.
 The low-precision parameters were then directly used in an event-based network of equivalent architecture.

 \subsection{Fast and efficient classification}

 \begin{figure}[t!]
     \centering
     \includegraphics[scale=0.9]{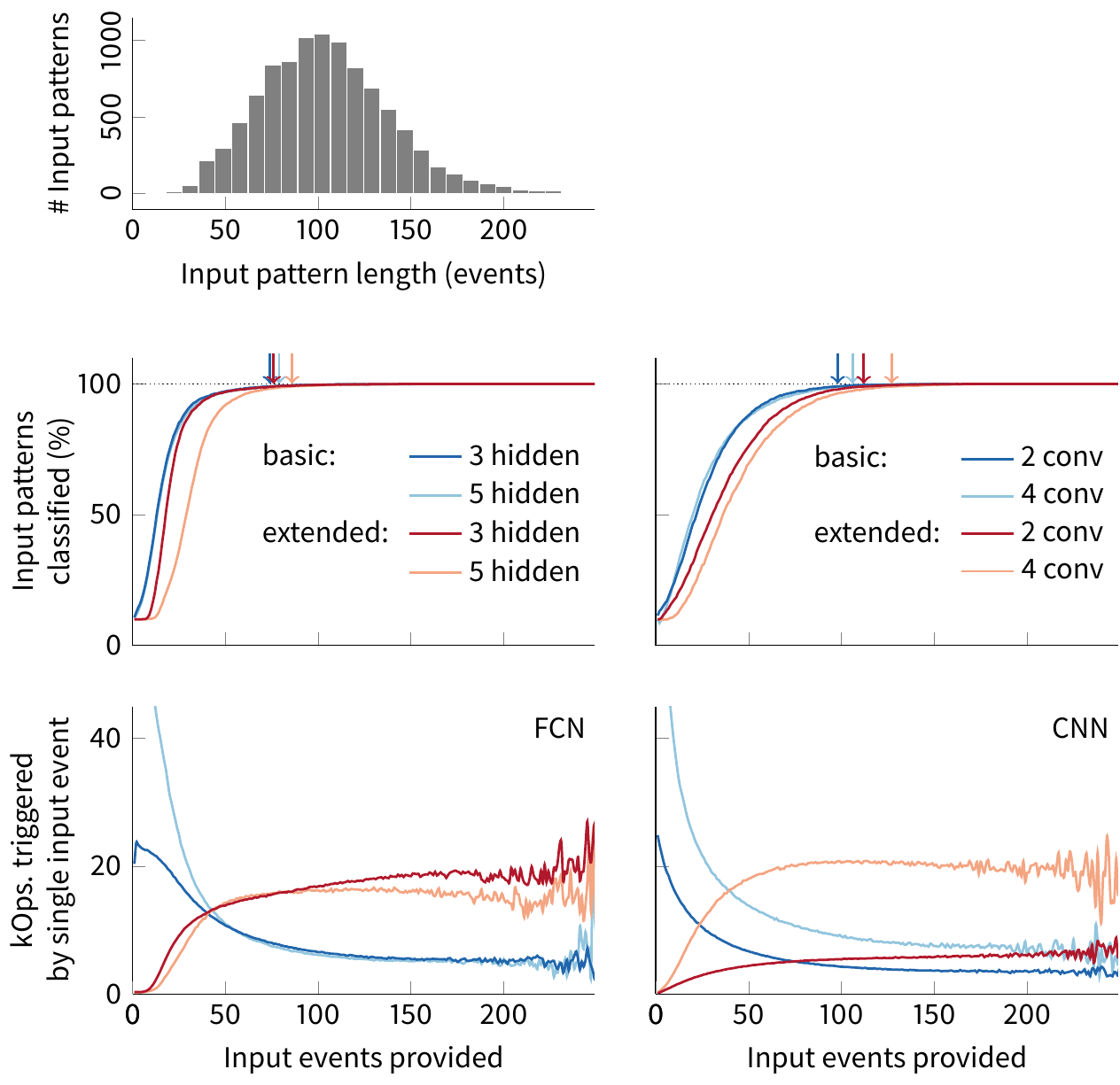}
     \caption{Performance of the basic and extended counter neuron models. The top panel shows the distribution of input pattern lengths (event count or active pixels) of the \mnist test set.
         The middle panels show the fraction of classified patterns (classified here means that the output is the same as that of the corresponding frame-based network) as a function of the number of input events presented to the network for the different architectures that were tested (mean over \mnist test set; left: FCNs; right: CNNs). The arrows mark the positions where the 99\,\% threshold is crossed.
         The bottom panels show the number of addition operations triggered in the networks by an individual input event over the course of the input presentation (mean over \mnist test set).
     }
     \label{fig:fcn-res}
 \end{figure}

\begin{figure}[t]
     \centering
     \includegraphics[scale=0.9]{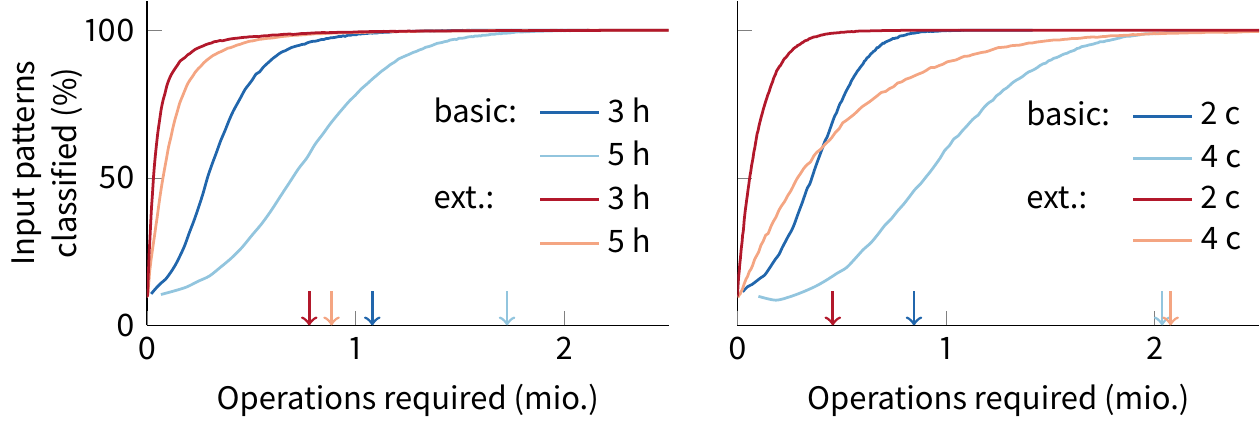}
     \caption{ Efficiency of the basic and extended counter neuron models.
	     The diagram shows the average number of operations required by different architectures to classify a certain fraction of test set examples. The arrows mark the positions where the 99\,\% threshold is crossed.
	     The color and line style correspond to the ones used in \cref{fig:fcn-res}.
     }
     \label{fig:fcn-ops}
 \end{figure}

 The main advantage of event-based deep networks is that outputs can be obtained fast and efficiently.
 We quantify this in \Cref{fig:fcn-res}, where the processing speed is measured as the time taken to produce the same output as the frame-based model, and efficiency is quantified as the number of addition operations required to compute the output. 
 For this analysis, individual pixels of the input image are provided to the network one by one in random order.
 In the systems based on the basic counter network model with binary neurons, the majority of events is emitted at the beginning of the input data presentation, with activity gradually declining in the course of the presentation.
 This reflects the fact that counter neurons cannot emit two events of the same sign in a row, leading to quick saturation in the output.
 The opposite is the case for the extended counter neuron based on the discretized ReLU, where activity gradually ramps up.
 Overall, this allows the extended model to produce the desired output with fewer operations than the basic model, as can be seen in \cref{fig:fcn-ops}.
 The achieved efficiency is beyond what had been possible with previous conversion-based methods: our method  achieves classification of \mnist at $<$500k events (CNN based on the extended neuron model), while the best reported result of a conversion-based network, to our knowledge, is $>$1 mio. events \cite{neil2016learning}.
 Despite the different network architectures, the behavior of FCNs and CNNs is qualitatively similar, with the main differences being due to the neuron model.
 In general, deeper networks seem to require a greater number of operations than shallower networks to achieve equivalent results.

 \section{Discussion}

 The two presented counter neuron models allow efficient event-based implementations of deep neural network architectures.
 While previous methods constructed deep spiking networks by converting parameters and approximating activations with firing rates, the output of our model is provably equivalent to its frame-based counterpart. Training is done in the frame-based domain, where state-of-the-art neural network optimization methods can be exploited.
 The discrete nature of counter networks allows hardware-friendly digital implementations, and makes them very suitable to process data from event-based sensors \cite{posch2014retinomorphic}.
 The resulting systems differ from traditional neural networks in the sense that units are updated `depth first', rather than `breadth first', meaning that any neuron can fire when its threshold is crossed, instead of waiting for all neurons in previous layers to be updated, as in conventional neural networks.
 This allows processing of input data as they arrive, rather than waiting for a whole frame to be transferred to the input layer. This can significantly speed up processing in digital applications.
 Compared to other deep spiking neural networks based on parameter conversion \cite{neil2016learning}, counter networks require fewer operations to process input images, even in our non-optimized setting. We expect that adding further constraints to enforce sparsity or reduce neuron activations can make counter networks even more efficient.
 Further research is required to investigate the applicability of the counter neuron model in recurrent networks.
 Finally, event-based systems are appealing because they allow for purely local, event-based weight update rules, such as spike-timing dependent plasticity (STDP).
 Preliminary results indicate that STDP-based training of counter networks is possible, which not only would allow efficient inference but also training of deep event-based networks.

 \subsection*{Acknowledgements}
 The research was supported by the Swiss National Science Foundation Grant 200021-146608 and the European Union ERC Grant 257219.

 \small
 \bibliographystyle{plain}
 \bibliography{bibliography}

\end{document}